# Privacy-preserving federated prediction of health outcomes using multi-center survey data

Supratim Das[a,b], Mahdie Rafie[a,b], Paula Kammer[a], Søren T. Skou[c,d], Dorte T. Grønne[c,d], Ewa M. Roos[c], André Hajek[b], Hans-Helmut König[b], Md Shihab Ullah[a], Niklas Probul[a], Jan Baumbach[a,e], Linda Baumbach[b]

[a]Institute for Computational Systems Biology, University of Hamburg, Albert-Einstein-Ring 8-10, 22761, Hamburg, Germany
[b]Department of Health Economics and Health Services Research, University Medical Center Hamburg-Eppendorf, Martinistraße 52 Hamburg, 20246, Germany
[c]Center for Muscle and Joint Health, Department of Sports Science and Clinical Biomechanics, University of Southern Denmark, Campusvej 55, 5230 Odense, Denmark
[d]The Research and Implementation Unit PROgrez, Department of Physiotherapy and Occupational Therapy, Næstved-Slagelse-Ringsted Hospitals, Fælledvej 2C, 4200 Slagelse, Denmark
[e]Computational Biomedicine Lab, Department of Mathematics and Computer Science, University of Southern Denmark, Odense, Denmark

---

**Abstract:**

**Background:** Patient-reported survey data are used to train prognostic models aimed at improving healthcare. However, such data are typically available multi-centric and, for privacy reasons, cannot easily be centralized in one data repository. Models trained locally are less accurate, robust, and generalizable. We present and apply privacy-preserving federated machine learning techniques for prognostic model building, where local survey data never leaves the legally safe harbors of the medical centers.

**Methods:** We used centralized, local, and federated learning techniques on two healthcare datasets (*GLA:D*® data from the five health regions of Denmark and international SHARE data of 27 countries) to predict two different health outcomes. We compared linear regression, random forest regression, and random forest classification models trained on local data with those trained on the entire data in a centralized and in a federated fashion.

**Results:** In GLA:D® data, federated linear regression (R2 0.34, RMSE 18.2) and federated random forest regression (R2 0.34, RMSE 18.3) models outperform their local counterparts (i.e., R2 0.32, RMSE 18.6, R2 0.30, RMSE 18.8) with statistical significance. We also found that centralized models (R2 0.34, RMSE 18.2, R2 0.32, RMSE 18.5, respectively) did not perform significantly better than the federated models. In SHARE, the federated model (AC 0.78, AUROC: 0.71) and centralized model (AC 0.84, AUROC: 0.66) perform significantly better than the local models (AC: 0.74, AUROC: 0.69).

**Conclusion:** Federated learning enables the training of prognostic models from multi-center surveys without compromising privacy and with only minimal or no compromise regarding model performance.

---

## 1. Introduction

Patient-reported survey data has emerged as a valuable tool for assessing medical conditions in clinical settings, serving as a gold standard in various fields[1–4]. Patient-reported survey data and objective functional tests have been instrumental in developing prognostic models using machine learning techniques[5,6]. Beyond its applications in personalized medicine, patient-reported survey data is also utilized in health economics research.

Clinical survey data is collected and stored across local hospitals and practices and sometimes across regions, states, or countries. Local data from a single silo is by nature smaller and hence leads to reduced accuracy and generalizability of prognostic statistical models as compared to models trained on larger data sets integrated from several clinics, regions, or countries[7,8]. The prime reasons for data decentralization are data security and privacy policies. Healthcare data exchange across country borders is particularly challenging[9]. Hence, real-world clinical survey data is scarce and rarely available for machine learning. In recent years, the idea of a decentralized electronic health record repository network across Europe has been suggested (but not yet addressed or implemented) in the framework of the European Union's proposal of a European Health Data Space (EHDS).[10]

### 1.1. Privacy regulations

Different privacy regulations around the world[11–15] effectively prohibit companies and researchers from directly sharing personal patient data. Despite patient ID anonymization attempts[16], recent studies demonstrate how various reidentification techniques may compromise patient privacy[17,18]. In a recent paper from 2022, Jill Evans *et al.* have pointed out the challenges in directly sharing osteoarthritis data; besides privacy, they mentioned logistical and structural barriers, such as the limitation of resources for data governance[19].

### 1.2. Consequence of scattered data silos in medicine

As a result, researchers are often constrained to train prognostic models solely based on local data, leading to suboptimal accuracy and potentially incorrect predictions (see Figure 1 for an illustration). In particular, as local data, in addition to its smaller size, possesses smaller sample diversity, the resulting prognostic machine learning models suffer from reduced generalizability and lead to increased discrimination of underrepresented minorities. Although a model trained on data from one clinical site might learn to predict with high accuracy for that particular clinic, the model might learn from some clinical site-specific information irrelevant to the disease in general[7,8]. Employing data from multiple hospitals for model training instead of relying solely on more homogeneous single-clinical data has been demonstrated to reduce model bias. Consequently, it leads to improved performance across the whole target population[7,20]. Besides this, even in scenarios where data centralization is possible, researchers need to go through many legal procedures and would need to take care of data storage logistics[19]; federated learning cut through these procedures, making data from several locations easy to use.

### 1.3. Federated learning

To circumvent legal and ethical data-sharing issues while allowing for the training of robust models on big data, researchers have developed federated learning[21,22] over the last few years. In federated learning, raw data can stay in the legally safe harbors of the local data centers. Instead, one trains a joint model using inference from all data distributed across the different hospitals (see Figure 1 for an illustration of the concept). For instance, federated learning has been used in computer vision[23] and natural language processing[24]. The efficacy of federated learning has been studied on image data[23], multi-omics data[25,26], and other biomedically relevant data types[22,27,28]. The applicability of a federated model in such data is often determined based on its ability to outperform local models[29,30] while maintaining a close resemblance to centralized models[29–31]. However, its potential and effectiveness for decentralized profiling of patient-reported survey data have been unexplored hitherto. The nature of survey data differs significantly from data obtained through genomic analyses, standard biochemical assays, or radiographic imaging. Unlike the continuous data typically encountered in these domains, survey data is often in the form of ordinal or nominal data. While ordinal data may be approximated as continuous under certain circumstances, it deviates from the characteristics of true continuous data[32].

Here, we aim to fill this gap by investigating the applicability of federated learning to patient-reported survey data. First, we predicted the development of self-reported pain in patients with osteoarthritis using data from all five regions of Denmark, based on previous work utilizing linear and random forest regression models[5]. Second, we use random forest classification models to predict physical inactivity using data from 27 European countries.

We anticipate that despite the distinct nature of our dataset, federated learning, consistent with previous studies, exhibits superior accuracy compared to local models across all scenarios[33]. Similarly, based on findings from existing studies, we anticipate comparable performance to the centralized counterparts[33,34], for the federated linear regression, random forest regression, and random forest classification.

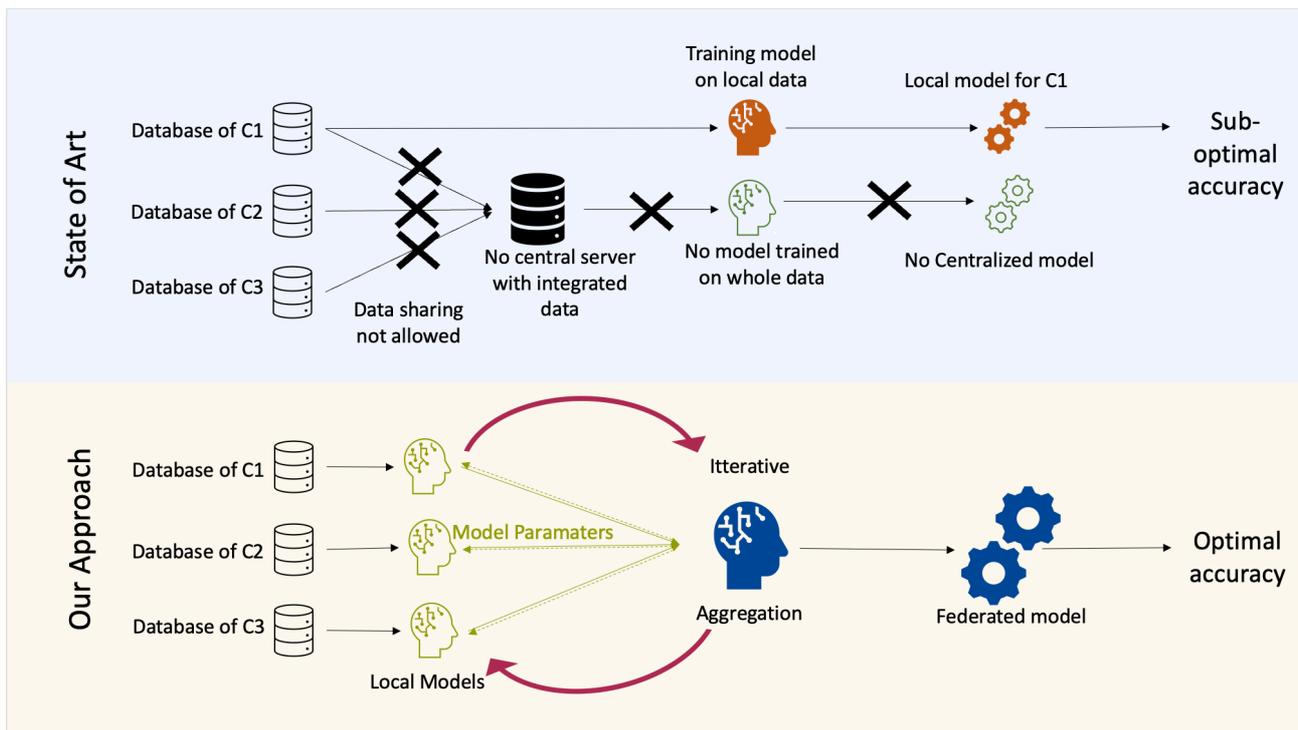

Figure 1: **Comparison of traditional machine learning and federated learning:** On the top half in the azure background, we represent the current state of the art of model predictions. In traditional machine learning, all data from data centers c1, c2, and c3 must be integrated into one server. Due to privacy issues and other logistic issues, integrating data from all local data centers to one central data site may not be possible. Hence, these traditional machine learning studies are often limited to training individual local models on limited data. These models have suboptimal performance as they fail to generalize in a global scenario. In the bottom part, in the ivory background, we depict our proposed model of federated learning. Here, the raw data never leaves their local servers c1, c2, and c3, and the necessity of data integration in one data center is mitigated. Each data center trains a local model; only model-specific parameters are communicated during training. As the federated machine learning model is trained using insights from all datasets, it can generalize better and potentially give better performance than a local model.

## 2. Method

### 2.1. Data and Preprocessing

For the first task, we utilized data from the *GLA:D®* (Good Life with osteoArthritis in Denmark) registry [35,36]. National GLA:D® databases exist in Australia, Austria, Canada, China, Germany, Ireland, Netherlands, New Zealand, and Switzerland[35–38]. The *GLA:D®* initiative consists of three parts: a certification course for physiotherapists, a standardized and supervised patient education and an exercise therapy program, and a registry with data from before, immediately after, and 12 months after program initiation[35–38]. *GLA:D®* aims to facilitate the implementation of recommended first-line pain management (exercise therapy and patient education) for patients with hip and knee osteoarthritis in clinical practice[39].

We utilized Danish data from the beginning of 2013 to the end of 2018. We used the same preprocessing described in a paper on individualized predictions of changes in knee pain for patients with knee osteoarthritis using the same data source[5], which includes a selection of the same 51 variables and dropping of patient data samples with missing values. Thus, we only included complete cases. To simulate a realistic federated learning scenario, we then split the data according to the five Danish administrative health care regions (North Denmark, Central Denmark, Southern Denmark, Zealand region, and the Capital Region of Denmark) using the geographical location where the (n=274) *GLA:D®* clinics are located[38]. For this, we call this task a national scenario.

We trained statistical models to predict changes in knee pain intensity over the period of approximately three months from before to after the *GLA:D®* program. We utilized the same 51 variables as Baumbach *et al.*[5] suggested. A detailed description of these 51 variables, which provides information about the patient's demographics and health status, is provided in supplementary table B1. Knee/hip pain intensity was measured on a VAS scale of 0 to 100 mm, best to worst. Hence, the possible values for pain change are between -100 and 100. The script used for preprocessing can be accessed in our GitHub repository[40].

For the second task, we used the data from wave-8 (released in 2022) of the Survey of Health, Ageing, and Retirement in Europe (SHARE) database[41,42] to predict physical inactivity. We chose this dataset because of the high heterogeneity in physical activity across 27 countries [43]. Hence, we expected this to impose greater challenges on the federated model. We calculated our outcome variable physical inactivity (yes/no) from two questions as suggested by Gomes et al. [43]. Afterwards, we selected 30 relevant variables, including information on mental health (such as EURO-D measure of depressive symptom[44,45]), cognitive abilities, movement limitations, self-perceived health, which are shown to be relevant factors in the literature[43] [see supplementary Table 3.] To simulate federated learning across multiple countries, we split the data into 27 countries. A full list of these variables and data preprocessing codes can be found in our GitHub repository[40].

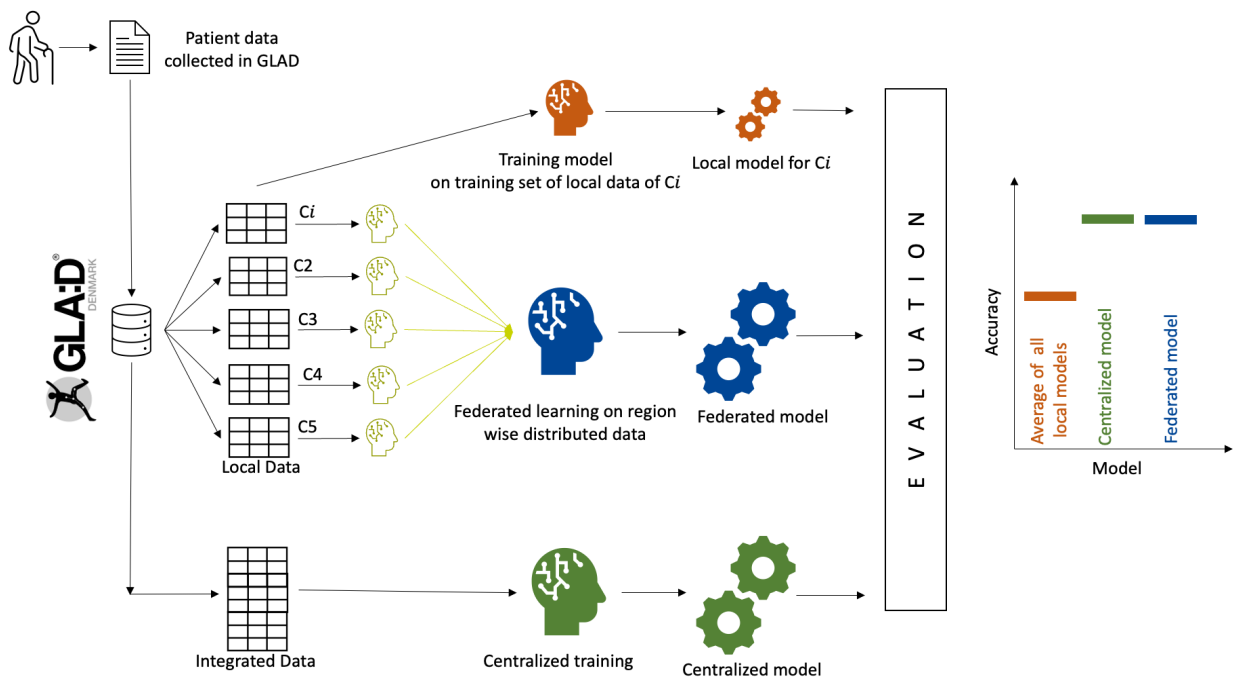

Figure 2: **Representation of the evaluation scheme for Federated Learning, exemplarily depicted for five regions of *GLA:D®* Denmark's data**: To compare with a centralized scenario, we simulate a scenario of distributed and centralized data centers using Danish *GLA:D®* data. We trained five local models with each region's local data, one federated model with distributed data, and one centralized model with integrated data. We evaluated this protocol in VAS pain prediction from 51 variables using linear regression and random forest regression as our machine-learning models of choice. We compared the average of 5 local models and a centralized scenario with the federated model for both machine learning algorithms (linear regression and random forest). A similar scheme was executed with the SHARE data but for 27 countries using random forest classification.

### 2.1.1. Local vs. centralized vs. federated machine learning

We trained the machine learning algorithms for both tasks in three different scenarios:

1. Local models: For task one, five separate models were trained in isolation on the local *GLA:D®* data of the five regions of Denmark. Similarly, for task two, 27 separate models were trained in isolation on local data from the 27 countries participating in SHARE (wave-8). In task two for the classification task, before training the model for each country, the data was balanced (same number of positive and negative samples) by supersampling (sampling with replacement from the smaller class).

2. Centralized model: For task one, local *GLA:D®* data of the five training sets were merged and used together to train one centralized model while having access to the full data. For task two, country-wise supersampled balanced training data sets of SHARE data were combined to create the international, centralized training set.

3. Federated model: For both tasks we trained the federated model using the same data used for local training but while keeping the regional/national sets separated (see Figure 2).

Finally, we trained linear regression, random forest regression, and random forest classification models for each of the three scenarios and evaluated their performances.

*2.2. Training and test data generation*

For both tasks, we performed 5-fold cross-validation, which gives us an 80% sample for training and a 20% sample for testing the machine learning models in each fold (see Supplementary Figure A.2). For task one, for each fold, during training, the federated and local models utilized the train split of their local data, while for centralized training, the same training splits are concatenated to form a centralized training set. In each fold, we used the global test set to evaluate the performances of local, centralized, and federated models. We tested all scenarios on the same global test set (see Supplementary Figure A.2).

Similarly, for task two, we split the data according to 27 countries, and in each country, we made a training test split in each fold of cross-validation. For the training split, we balanced the dataset by supersampling. For evaluation, we first merged the local test sets into centralized ones; then, we created 4 subsampled balanced subsets where the sample of the larger class is mutually exclusive in the 4 subsets. We do this for fair evaluation on a balanced test set.

*2.3. Machine Learning Methods*

*2.3.1. Scenario 1: Local models*

For the local models, we used linear regression, random forest regression, and random forest classification models trained on a training split of the respective local data of each of the regions (for task one) or countries (for task two). Thus, we have five separate models for each algorithm in each fold. For all local models, we utilized the sci-kit-learn library[46] in Python.

*2.3.2. Scenario 2: Centralised model*

For the centralized models, we used the same sci-kit-learn library in Python to train linear regression, random forest regression, and random forest classification models. Here, for every machine learning model, only one model is trained in each fold, which utilizes the integrated training data (see Supplementary Figure A.2).

 *2.3.3. Scenario 3: Federated model*

In federated learning, instead of sharing raw data, each data center or clinic, also referred to as "client," locally performs computations on their local data and only shares model parameters with an aggregator[9,21–23,33]. The aggregator then combines all those parameters and sends them back to each client. Now, each client updates their machine learning model using these values. In iterative training schemes (like random forest classification, for instance), these two steps are iterated until the model converges. See Supplementary Figure A.1 for an illustration. Federated learning, in brief, works as follows:

1.      Initialization: All clients are initialized with the same model architecture; this might be created from a local configuration file or received from the coordinator, which can be an external or one of the participating clients. The coordinator can be the same as the aggregator.
2.      Local training: At this stage, each local client trains their local model using local data. For an iterative model, these local models are trained for $s$ local gradient descent steps before sending parameters for the communication step.
3.      Communication: In this step, each client sends local parameters or local gradients to the aggregator.
4.                                                                                          Global aggregation: The aggregator combines all the received local parameters from the clients and updates the shared model. This can be performed using simple weighted averaging.
5.      Convergence: Steps 2-4 are repeated until preset convergence criteria are met. The federated random forest classifier is implemented as an iterative model.
6.      Model evaluation**:** After training termination, each client saves the model, performs prediction on the local test set, and outputs local results for evaluation.
For federated learning, we used the feature cloud platform[33], which provides users with an interactive user interface to run commonly used machine learning algorithms in a federated fashion.

While federated linear regression can be implemented by sharing $X^{\top} \circ X$ and $X^{\top} \circ Y$ [47] where $X$ is an *n,m* matrix of *n* samples and *m* features and $Y$ is the target variable vector (which comes with considerable communication overhead), we exchange the slopes and intercepts[48] and calculate the weighted average of those. For the federated random forest regression models, local random forest regressor models were first trained, and the estimators of random forest or decision trees were gathered to create a new random forest model in the aggregation stage [33]. Unlike federated random forest regression, the random forest classification model is designed as an iterative, histogram-based model. To generate this model, a predetermined amount of identical threshold points are evaluated for each client, selected feature, and decision tree node. Doing this guarantees finding the optimal threshold point for an aggregated model in a privacy-preserving fashion[49]. This process is repeated for each tree to generate a globally optimal model. Detailed information on the utilized FeatureCloud app versions and the applied configuration files can be found in our GitHub repository[40].

## 2.3. Evaluation

For all the models, we first evaluated their performances using 5-fold cross-validation. In all three scenarios (local, centralized, and federated), we calculated R-squared and root mean squared error (RMSE) on the global test data to evaluate linear regression and random forest regression. For the Random forest classifiers in each fold of cross-validation, we tested 4 different subsampled unbiased centralized test sets for fairness. In those 4 test sets, the negative samples were mutually exclusive. For evaluation, we used both accuracy and AUROC.

We performed single-sided tests to check our hypotheses on whether centralized and federated models outperform the local and whether the centralized analyses perform better than federated ones. We performed the student's t-test when the assumptions of normality and equal distribution were met and the Welch t-test when at least one of those conditions was not met. The normality of distribution was checked using Shapiro's test, and we checked the homoscedasticity to verify the equal distribution assumption using Bartlette's test. Python scripts for evaluation and statistical tests are available through our GitHub repository[40].

## 2.3. Criteria for applicability of federated learning

Being among the first to apply federated learning to survey-based data, we used literature from computer vision and other biomedical data to define when federated learning would be applicable to survey data. In computer vision tasks, comparisons to local models are often omitted, and the threshold for determining federated algorithm performance relative to centralized models remains unspecified[50–52]. In federated learning literature, only an improvement over other federated algorithms is reported[53,54]; the statistical significance tests are often left out. The precise cutoff for the extent of similarity to centralized accuracy remains undefined in the literature and depends on the specific application and the trade-off between privacy and accuracy. In light of these findings, we adopt the following two criteria: if the federated model's accuracy surpasses that of local models and aligns closely with centralized model performance, we deem it applicable. Thus, a necessary criterion to deem federated learning applicable is that the federated models perform statistically significantly better than local models. However, if centralized models do not outperform federated models, it is a sufficient criterion.

## 2.4. Additional analyses

For the tasks based on the GLA:D® data, we performed some additional analysis to test if the regions are statistically significantly different from each other and how they perform in the data scares scenario.

### 2.4.1 Subsampling experiment

To investigate the efficacy of federated learning in scenarios where each client possesses limited data, we replicated our experiments based on the GLA:D® data with reduced training data. The training set size was progressively reduced to 75%, 50%, 25%, and 10% of its original size, while the central test set remained unchanged. We compared the performance of linear regression and random forest models trained locally and centrally using federated learning under these reduced data conditions. For the 25% subsampling, we employed a five-fold cross-validation procedure. We performed the same evaluation and statistical tests to produce results comparable to those obtained with the entire training set.

### 2.4.2. Data heterogeneity

We also tested the heterogeneity of all variables across all the regions, as data heterogeneity might specifically affect the performance of federated models. For all 51 variables and the target variable, we tested for each pair of regions to see whether the data was coming from two different distributions. To this end, we performed post-hoc analyses using the Scheffe score. To evaluate the extent of the effect size of the difference in distribution, we used Cohen's D.

## 3. Results

After preprocessing the *GLA:D®* data, we included 9,648 patients. The number of samples per region is given in Table 1.

For the SHARE dataset, we included 46570 patients from 27 countries. The number of samples per region varies between 2933 and 509 participants; details are provided in supplementary table 2.

| Table 1: **Distribution of patients in *GLA:D®* registry across five different regions of Denmark** | | |
|---|---|---|
| Region name | Region ID | number of patients |
| Capital Region of Denmark | 1 | 2415 |
| Region Zealand | 2 | 1550 |
| Region of Southern Denmark | 3 | 2530 |
| Central Denmark Region | 4 | 2220 |
| North Denmark Region | 5 | 933 |
| total | .. | 9648 |

*3.1. Performance evaluation:*

All results for comparisons of federated models with centralized and local models are given below in Table 2. The p-values for the statistical tests are given in Table 3.

| Table 2: **Comparison of Federated learning with centralized and local models when trained on the full *GLA:D®* dataset and wave 8 of the SHARE dataset.** | | | | | | |
|---|---|---|---|---|---|---|
| data/ scenario | Model | Metric | Average of Locals | Centralized | Federated | Relative Improvement of federated over local |
| *GLA:D®*, National | Linear Regression | r-square | 0.32 | 0.34 | 0.34 | 8.16% |
| *GLA:D®*, National | Linear Regression | RMSE | 18.6 | 18.2 | 18.2 | 2.09% |
| *GLA:D®*, National | Random Forest Regression | r-square | 0.30 | 0.32 | 0.34 | 8.16% |
| *GLA:D®*, National | Random Forest Regression | RMSE | 18.8 | 18.5 | 18.3 | 2.01% |
| SHARE, International | Random Forest Classification | accuracy | 0.74 | 0.84 | 0.78 | 5.45% |
| SHARE, International | Random Forest Classification | AUROC | 0.69 | 0.66 | 0.71 | 2.89% |

| Table 3: **Statistical tests for comparison of federated learning with centralized and local models when trained on the full *GLA:D®* dataset and wave 8 of the SHARE dataset.** | | | | | |
|---|---|---|---|---|---|
| data/ scenario | Model | Metric | If Centralised better than Local (P-value) | If Federated is better than Local (P-value) | If Centralised is better than Federated (P-value) |
| *GLA:D®*, National | Linear Regression | r-square | 0.00076 | 0.00092 | 0.48 (n.s.) |
| *GLA:D®*, National | Linear Regression | RMSE | 0.0390 | 0.040 | 0.49(n.s.) |
| *GLA:D®*, National | Random Forest | r-square | 0.012 | 0.00010 | 0.96(n.s.) |
| *GLA:D®*, National | Random Forest | RMSE | 0.10 (n.s.) | 0.017 | 0.75(n.s.) |
| SHARE, International | Random Forest Classification | accuracy | 3.5e-37 | 3.4e-12 | 1.2e-8 |

| SHARE, International | Random Forest Classification | AUROC | 0.99(n.s.) | 4.6e-7 | 1.0n.s.) |

n.s. : statistically not significant.

### 3.1.1. Linear Regression

For linear regression used in task 1, federated learning models have a higher R-squared and lower RMSE than the corresponding average of the local models (R-squared p-val 0.00092 RMSE p-val 0.040). While the federated model archives R-squared 0.34 and RMSE 18.2, the local models on average archives R-squared 0.32 and RMSE 18.6 (Table 2). Conversely, the centralized model archives R-squared 0.34 and RMSE 18.2 (Table 2). Hence, the centralized model does not outperform the federated model (R-squared p-val 0.48, RMSE p-val 0.49)(Table 3). Note that the local model trained on data from the North Denmark Region (Region 5) performs worst among the local models (Figure 3).

### 3.1.2. Random Forest Regression

For random forest regression used in task 1, federated learning models have a higher R-squared and lower RMSE than the corresponding average of the local models (R-squared p-val 0.00010 RMSE p-val 0.017). While the federated model archives R-squared 0.34 and RMSE 18.3, the local models on average archives R-squared 0.30 and RMSE 18.8 (Table 2). Conversely, the centralized model archives R-squared 0.32 and RMSE 18.5 (Table 2). Hence, the centralized model does not outperform the federated model (R-squared p-val 0.96, RMSE p-val 0.75)(Table 3). Rather, the federated model outperforms the centralized model (Figure 3).

### 3.1.3. Random Forest Classification

In random forest classification, used in task 2, the centralized model accuracy slightly outperforms(0.84) the federated model (0.78). Both models outperform the local model (0.74) (Supplementary Figure A.3.). However, the federated model achieved higher AUROC (0.71) than the centralized (0.66), and the average of local models ( 0.69) (Figure 3). A higher accuracy but lower AUROC in the centralized model over the federated model suggests that federated random forest classification is more robust and less prone to overfitting.

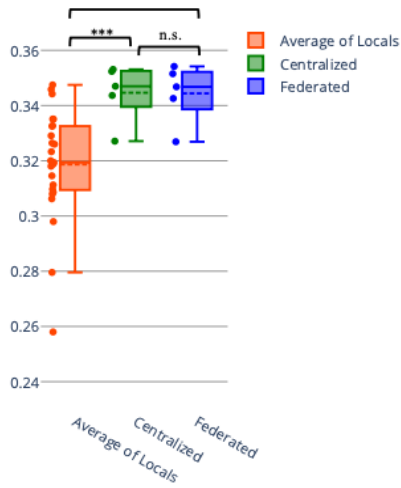

A.  Linear Regression – r- square

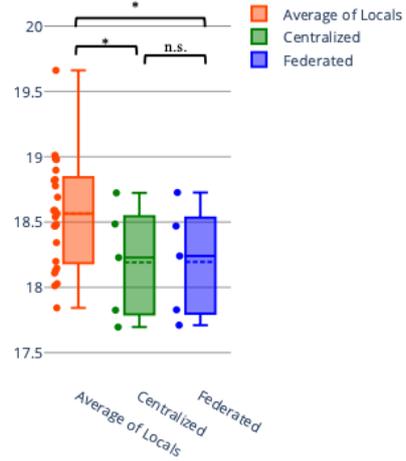

B.  Linear Regression – RMSE

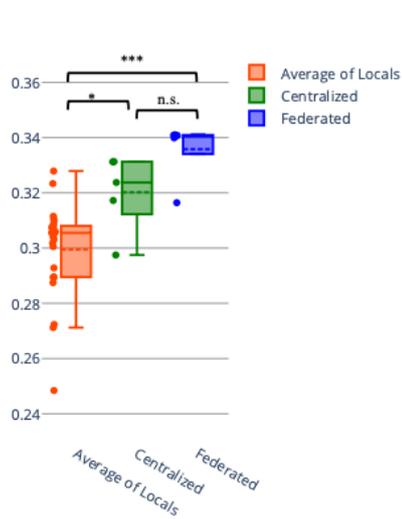

C.  Random Forest Regression – r- square

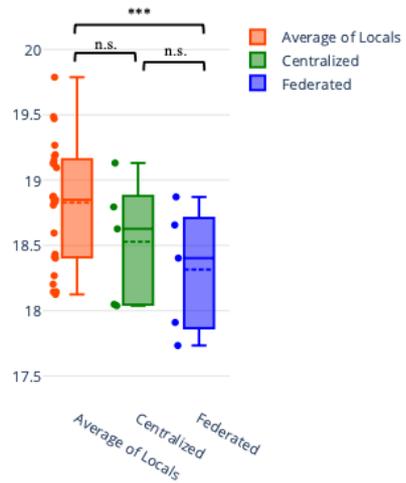

D.  Random Forest Regression – RMSE

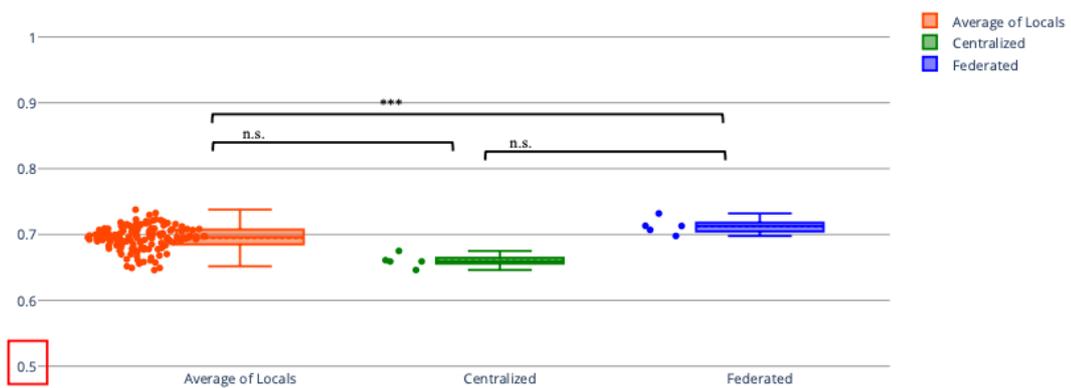

E.  Random Forest Classifier – AUROC

Figure 3: **Comparison of federated model with centralized and local models**: For both **A.** R-square of linear regression (higher the better) and **B.** RMSE plot of linear regression (lower the better), we see federated and centralized models performing better than the local models. Similarly, for both **C.** R-square of random forest regression (higher the better) and panel **D.** RMSE of random forest regression (lower the better), we see that the federated model outperforms both local and central models. In **E.** For the random forest classifier, we see that the federated model has higher AUROC than centralized and local. The dotted lines in the box plots represent the mean, while the solid lines represent the median. The stars represent the magnitude of the significance of the tested hypothesis in the following convention *** : p<0.001, ** : p<0.01, * : p<0.05, n.s. : p>0.05. We tested, 1. if the centralized model is better than the average of local ones, 2. if the federated is better than the average of local ones, and 3. if the centralized is better than the federated.

*3.2 Supplementary analyses*

*3.2.1 Subsampling experiment*

In the first task, the performance gap between local models and centralized models widens as the training dataset size diminishes (see Figure 4). Still, federated models manage to keep up with their centralized counterparts. In the 25% subsampled training set experiment (range n= 88  to n= 266), we observe a remarkable 55.3% improvement in R-squared for linear regression when using federated learning compared to the average performance of local models across five-fold cross-validation (Table 3 and Supplementary Figure A.5.). Federated random forest exhibits a substantial 20.9% improvement in R-squared value over its local counterpart. Unlike the experiment in full data in this data-scarce simulation, the statistical test always finds a statistically significant improvement of centralized and federated over local models. However, there is no significant improvement in centralized over the federated model (Table 5).

Table 4: **Comparison of Federated learning with centralized and local models when trained on only 25% of training data in *GLA:D®*.**

| Model trained on 25% of training data | Metric | Average of Locals | Centralized | Federated | Relative Improvement of federated over local |
|---|---|---|---|---|---|
| Linear Regression | r-square | 0.21 | 0.31 | 0.32 | 55.32% |
| Linear Regression | RMSE | 19.9 | 18.7 | 18.5 | 7.42% |
| Random Forest Regression | r-square | 0.27 | 0.31 | 0.32 | 20.83% |
| Random Forest Regression | RMSE | 19.2 | 18.6 | 18.5 | 3.89% |

Table 5: **Statistical tests for Comparison of Federated learning with centralized and local models when trained on only 25% of training data in *GLA:D®*.**

| Model | Metric | If Centralised better than Local (P-value) | If Federated is better than Local (P-value) | If Centralised is better than Federated (P-value) |
|---|---|---|---|---|
| Linear Regression | r-square | 1.23E-06 | 1.80E-07 | 0.95(n.s.) |
| Linear Regression | RMSE | 0.00067 | 1.01E-05 | 0.75(n.s.) |
| Random Forest Regression | r-square | 0.00045 | 1.35E-08 | 0.98(n.s.) |
| Random Forest Regression | RMSE | 0.021 | 3.39E-03 | 0.73(n.s.) |

n.s. : statistically not significant.

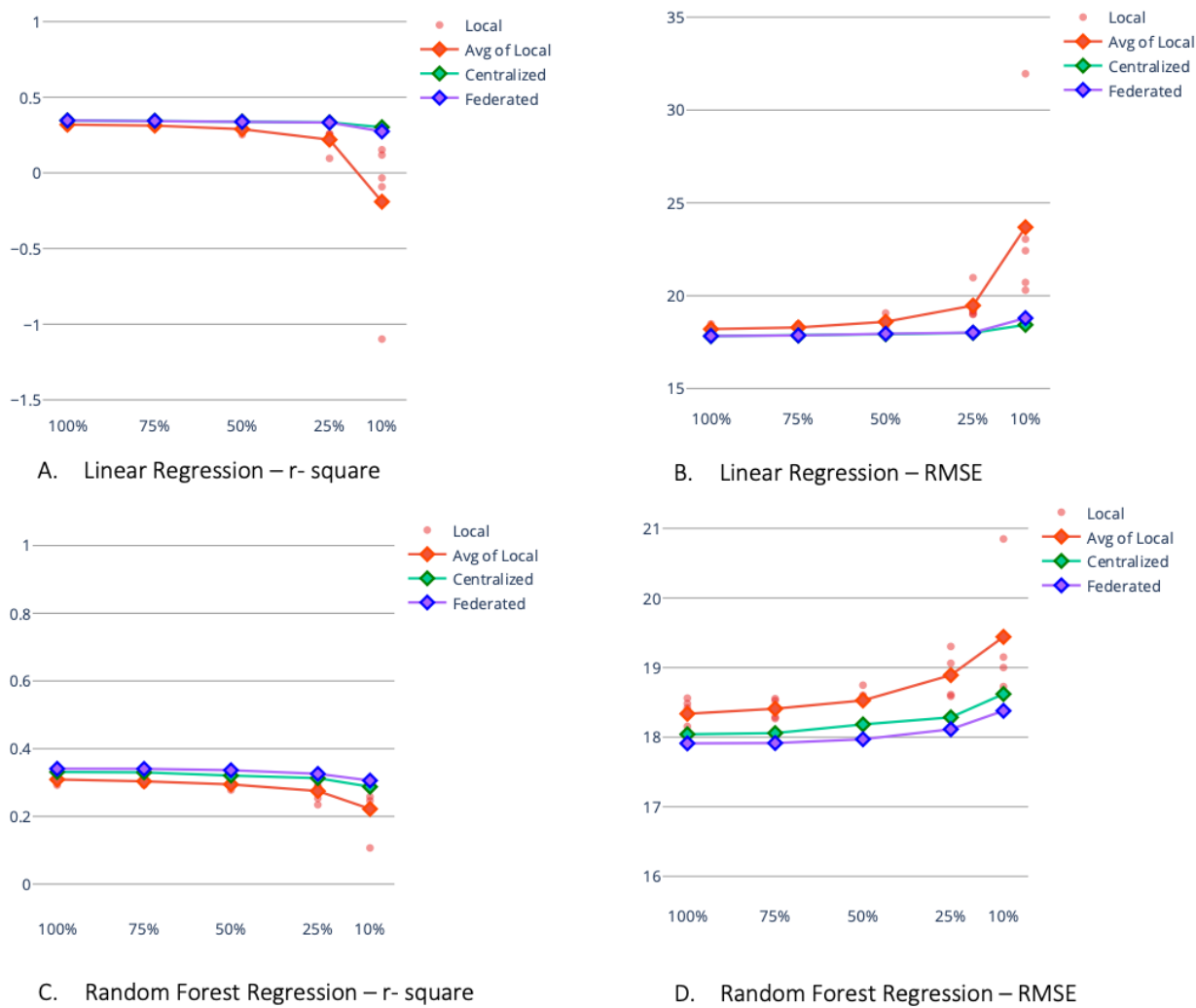

Figure 4: **Error rates after training set subsampling experiment in *GLA:D*®**: The training data has been subsampled to 75%, 50%, 25%, and 10%, while test data is kept untouched. Linear regression, and random forest regression are trained on subsampled *GLA:D*® data to compare federated models with centralized and local models. For both machine learning experiments, although the local model, on average, performs worse than the federated model, the difference between them intensifies with a reduction in the sample size of training data.

### 3.2.2 Data heterogeneity

For the national scenario, in Danish *GLA:D*® data, although the means of the distribution of each region per each variable are very close and most groups do not differ statistically significantly, we observed some statistically significant differences (see Figure 5(a) and supplementary Figure A.5(a)) in some of the variables (age, change in VAS pain score). In terms of multiple variables, region five (North Denmark Region) is significantly different (lighter color in Figure 5(a) and Supplementary Figure A.5(a) for region 1 vs. 5, 2 vs. 5, 3 vs. 5, 4 vs. 5). We also tested the Cohen's D effect size to determine what population of one group is below the average of the mean of the other group. Our results suggest that although most of them have small effect sizes, a few of them (e.g., age, change in VAS pain score) have larger effect sizes (Figure 5(b) and Supplementary Figure A.5 (b)). Cohen's D effect size concurs with the post hoc Scheffe analysis (Figure 5 and Supplementary Figure A.5).

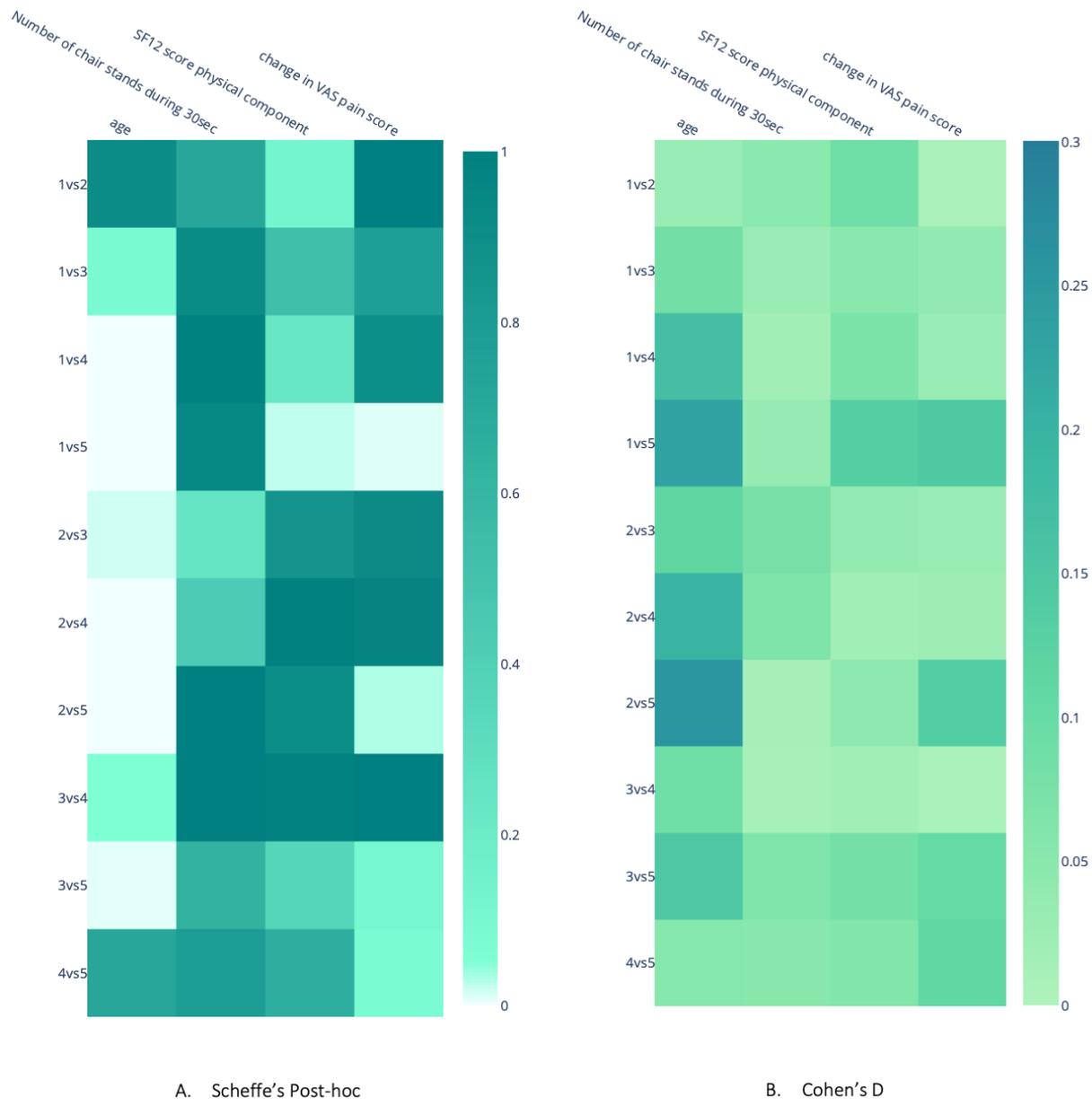

A.  Scheffe's Post-hoc                                    B.  Cohen's D

Figure 5: **Statistical test to evaluate data heterogeneity of *GLA:D®* across five regions**: **A. Posthoc Scheffe test:** We used post hoc Scheffe to test pairwise if two regions' data distributions differ. The comparison of two regions, X and Y, is denoted as "XvsY." We did this for all 51 variables and the target variable. Here, we plot this for only 3 of the variables: age, a functional test of the number of chair stands, SF12 score, and our target variable: Change in VAS pain score. The light color suggests statistical significance, while the darker color suggests no significance. **B. Cohen's D** effect size of each statistical test.: the darker the color, the higher the effect size. Although the means of the distribution are close, we find statistically significant differences among regions in some variables, including the target variable. Find detailed post hoc Scheffe and Cohen's D analysis plots of all variables in the supplementary material.

## 4. Discussion

 Following the set criteria for the applicability of federated learning, our findings indicate that federated learning applied to patient-reported survey data outperforms models trained on local data with statistical significance, while the centralized models are comparable to the federated model.

In federated learning, sometimes a compromise in accuracy is acceptable for privacy. For example, a compromise of approximately 5% accuracy in brain tumor prediction from MRI images is considered acceptable in federated convolutional neural networks for privacy preservation[31]. However, our federated models in *GLA:D®* data never underperform the centralized models. In SHARE data, although the federated random forest classifier had lower accuracy than centralized, it has higher AUROC than centralized and, hence, is more robust.

Across all three machine learning models investigated federated models exhibited higher accuracy compared to their local counterparts, as expected. Federated linear regression achieved performance identical to that of its centralized counterpart (same R-squared and accuracy, respectively). Remarkably, the federated random forest regressor in task 1 (R-squared = 0.34) and federated random forest classifier in task 2 (AUROC = 0.71) even surpassed the performance of the centralized counterparts (0.32, 0.66 respectively), a rare phenomenon that has also been observed in multiple biological data[33]. We speculate this to happen due to the overfitting of the centralized random forest regression model. We find for task 2; the centralized random-forest classifier model overfits, resulting in higher accuracy (0.84) and lower AUROC (0.66) than the federated model (AC 0.78, AUROC 0.71). This suggests that the federated random forest classification model, although it has slightly lower accuracy, is more robust.

We also sometimes see a few local models having very good performance against centralized test sets while most other local models underperform. However, in real life it would not be possible to find out the best local model against centralised test sets. In our simulations, we had the "luxury" of comparing different local models against a centralized ground truth by experimental design. In a real-world scenario, this would be impossible.

Our experiment in international data suggests the applicability of federated learning algorithms, even when each county has a different distribution in the target variable (see Supplementary Figure A.6.).

Overall, we see a clear benefit of using federated learning in patient-reported survey data when gathering local data into one server is impossible or if the data are scarce, as demonstrated in our subsampling experiment. As the sample size diminishes, the gap between local and centralized widens drastically while the federated keeps up with the centralized model. We also see that the local models are much more vulnerable to data diversity in international scenarios, while the federated model is robust to such data diversity.

Essentially, we demonstrated that for analyzing health survey data one does not need to risk data privacy violations by centralizing patient-reported survey data in one server to secure model performances. Decentralized databases, according to the European Commission's draft for the European Health Data Space (EHDS), can be materialized without affecting machine-learning-based prognostic model building for standardized distributed patient-reported survey data or health registries such as the *GLA:D®* and SHARE registry. Whenever centralization of local data is not possible, federated learning could be applied for prognostic model building, preferably even if one data center has sufficient data for local model training, as these local models only reflect their region's local population distributions. In contrast, federated learning across multiple regions trains more generalized models for the global population.

We further noticed that the local model originating from one of the five regions underperforms the other local models, which could be explained by the statistically significant difference in the distribution of the target variable (change in pain) and other variables (age, BMI, living situation) between region five and the other four regions. This illustrates that locally trained models may suffer heavily from data heterogeneity in real-world scenarios, where federated learning is less susceptible.

In our study, for the first time, we demonstrate the applicability of federated learning in real-life patient-reported survey data using federated linear regression, random forest regression, and random forest classification. The only requirement for federated learning is the availability of the same variables and target variables across all clients. Hence, whenever centralization of all isolated data is impossible or inconvenient, we suggest using federated learning to increase model performance by leveraging multi-center data, e.g., for developing personalized healthcare prediction models.

## 5. Conclusion

We evaluated the utility of federated learning technology and its applicability to patient-reported survey data for developing prediction models in healthcare research across nationwide and international scenarios. We demonstrated the applicability of federated machine learning in real-world datasets of patient-reported survey-based data. In both the national and international datasets, we found the federated models comparable to those trained in a centralized fashion while performing better than models trained on local data. Henceforth, we showed that federated learning can be applied to handle the privacy vs. model accuracy dilemma in multi-centric internationally distributed patient-reported healthcare datasets.

## 6. Code, Results, and Data Availability

All codes for data preprocessing, local and centralized models, evaluation, and details of repositories of the exact version of the feature cloud app can be accessed through our GitHub repository[40]. We have also added an HTML file for interactive plots where one can hover over the plots to see exact values in the same repository. For anonymized raw data access, one can apply to the co-leads of the *GLA:D®* Denmark's registry directly (Prof. Roos and Prof. Skou), who will evaluate requests respecting the General Data Protection Regulation (GDPR). For SHARE data, one can apply it directly to their online portal[41].

## 7. Conflict of interest

Dr. Roos is the copyright holder of Knee Injury and Osteoarthritis Outcome Score (KOOS) and several other patient-reported outcome measures and co-founder of Good Life with Osteoarthritis in Denmark (*GLA:D®*), a not-for-profit initiative hosted at the University of Southern Denmark aimed at implementing clinical guidelines for osteoarthritis in clinical practice. Dr. Skou has received personal fees from Munksgaard, TrustMe-Ed, and Nestlé Health Science, all of which are outside the submitted work. He is a co-founder of *GLA:D®*. Prof. Jan Baumbach is co-owner of the for-profit Dehaze GmbH, a company for machine learning solutions in biomedicine.

## 8. Funding Source:

This work was developed as part of the PhysioAI project and is funded by the German Federal Ministry of Education and Research (BMBF) and by the NextGenerationEU Fund of the European Union under grant number 16DKWN115A. The responsibility for the content of this publication lies with the author.

This project has also received funding from the European Union's Horizon2020 research and innovation program under grant agreement No 826078. This publication reflects only the authors' view, and the European Commission is not responsible for any use that may be made of the information it contains.

**9. Acknowledgment:** The authors would like to thank the clinicians and patients involved in collecting data for the *GLA:D®* registry. This paper uses data from SHARE Waves 8(DOIs: 10.6103/SHARE.w8.900,) see Börsch-Supan et al. (2013) for methodological details.(1) The SHARE data collection has been funded by the European Commission, DG RTD through FP5 (QLK6-CT-2001-00360), FP6 (SHARE-I3: RII-CT-2006-062193, COMPARE: CIT5-CT-2005-028857, SHARELIFE: CIT4-CT-2006-028812), FP7 (SHARE-PREP: GA N°211909, SHARE-LEAP: GA N°227822, SHARE M4: GA N°261982, DASISH: GA N°283646) and Horizon 2020 (SHARE-DEV3: GA N°676536, SHARE-COHESION: GA N°870628, SERISS: GA N°654221, SSHOC: GA N°823782, SHARE-COVID19: GA N°101015924) and by DG Employment, Social Affairs & Inclusion through VS 2015/0195, VS 2016/0135, VS 2018/0285, VS 2019/0332, VS 2020/0313 and SHARE-EUCOV: GA N°101052589 and EUCOVII: GA N°101102412. Additional funding from the German Ministry of Education and Research, the Max Planck Society for the Advancement of Science, the U.S. National Institute on Aging (U01_AG09740-13S2, P01_AG005842, P01_AG08291, P30_AG12815, R21_AG025169, Y1-AG-4553-01, IAG_BSR06-11, OGHA_04-064, BSR12-04, R01_AG052527-02, HHSN271201300071C, RAG052527A) and from various national funding sources is gratefully acknowledged (see www.share-eric.eu).

**Privacy-preserving federated prediction of pain intensity change based on multi-center survey data**

Supratim Das[a,b], Mahdie Rafie[a,b], Paula Kammer[a], Søren T. Skou[c,d], Dorte T. Grønne[c,d], Ewa M. Roos[c], André Hajek[b], Hans-Helmut König[b], Md Shihab Ullah[a], Niklas Probul[a], Jan Baumbach[a,e], Linda Baumbach[b]

[a]*Institute for Computational Systems Biology, University of Hamburg, Albert-Einstein-Ring 8-10, 22761, Hamburg, Germany*
[b]*Department of Health Economics and Health Services Research, University Medical Center Hamburg-Eppendorf, Martinistraße 52 Hamburg, 20246, Germany*
[c]*Center for Muscle and Joint Health, Department of Sports Science and Clinical Biomechanics, University of Southern Denmark, Campusvej 55, 5230 Odense, Denmark*
[d]*The Research and Implementation Unit PROgrez, Department of Physiotherapy and Occupational Therapy, Næstved-Slagelse-Ringsted Hospitals, Fælledvej 2C, 4200 Slagelse, Denmark*
[e]*Computational Biomedicine Lab, Department of Mathematics and Computer Science, University of Southern Denmark, Odense, Denmark*

## Supplementary Text:

Statistical Test details: For the algorithm, we will denote the results of the average of Locals as L , centralized as C, and federated model as F.

The algorithm for statistical tests is as follows:

```
| If for the metric higher is better : (e.g. R-squared, AUROC, Accuracy)
|        | Calculate the p-values for normality of L, C, and F using Shapiro's test.
|        | Calculate the p-values for test equality of distribution for L vs. C, L vs. F, and C vs. F using
|        Bartlett's test.
|        | If the Shapiro test's p-value for L, C, and p-value for equality of distribution of L vs C is >0.05:
|        |        | Calculate students' t-test, alternate hypothesis C>L
|        |else: calculate Welch's t-test, alternate hypothesis C>L
|
|        | If the Shapiro test's p-value for L, F, and p-value for equality of distribution of L vs F is >0.05:
|        |        | Calculate students' t-test, alternate hypothesis F>L
|        |else: calculate Welch's t-test, alternate hypothesis F>L
|
|        | If the Shapiro test's p-value for C, F, and p-value for equality of distribution of C vs F is >0.05:
|        |        | Calculate students' t-test, alternate hypothesis C>F
|        |else: calculate Welch's t-test, alternate hypothesis C>F
|
| If the metric lower is better (e.g. RMSE)
|        | Calculate the p-values for normality of L,C, and F using Shapiro's test.
|        | Calculate the p-values for test equality of distribution for L vs. C, L vs. F, and C vs. F using
|        Bartlett's test.
|        | If the Shapiro test's p-value for L, C, and p-value for equality of distribution of L vs C is >0.05:
|        |        | Calculate students' t-test, alternate hypothesis C<L
|        |else: calculate Welch's t-test, alternate hypothesis C<L
|
```

```
|       | If the Shapiro test's p-value for L, F, and p-value for equality of distribution of L vs F is >0.05:
|       |       | Calculate students' t-test, alternate hypothesis F<L
|else: calculate Welch's t-test, alternate hypothesis F<L
|
|       | If the Shapiro test's p-value for C, F, and p-value for equality of distribution of C vs F is >0.05:
|       |       | Calculate students' t-test, alternate hypothesis C<F
|else: calculate Welch's t-test, alternate hypothesis C<F
```

## Supplementary Figures

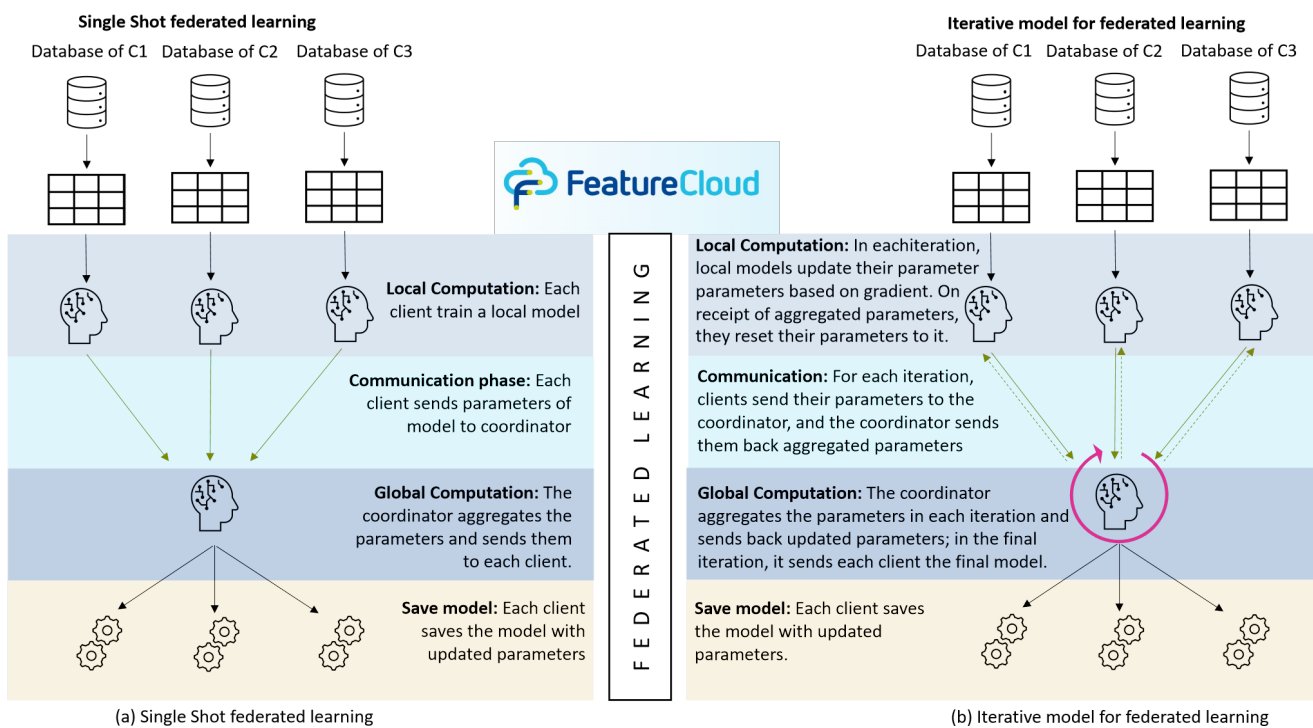

Supplementary Figure A.1: **Federated Learning in Feature Cloud, Modus operandi**: Federated learning of any centralized model can be achieved in two ways. One may use one-time communication of local models with the aggregator and update the aggregated parameters. Simpler models can be built in this manner. Another way is to have multiple such communication rounds (iterations) until convergence.

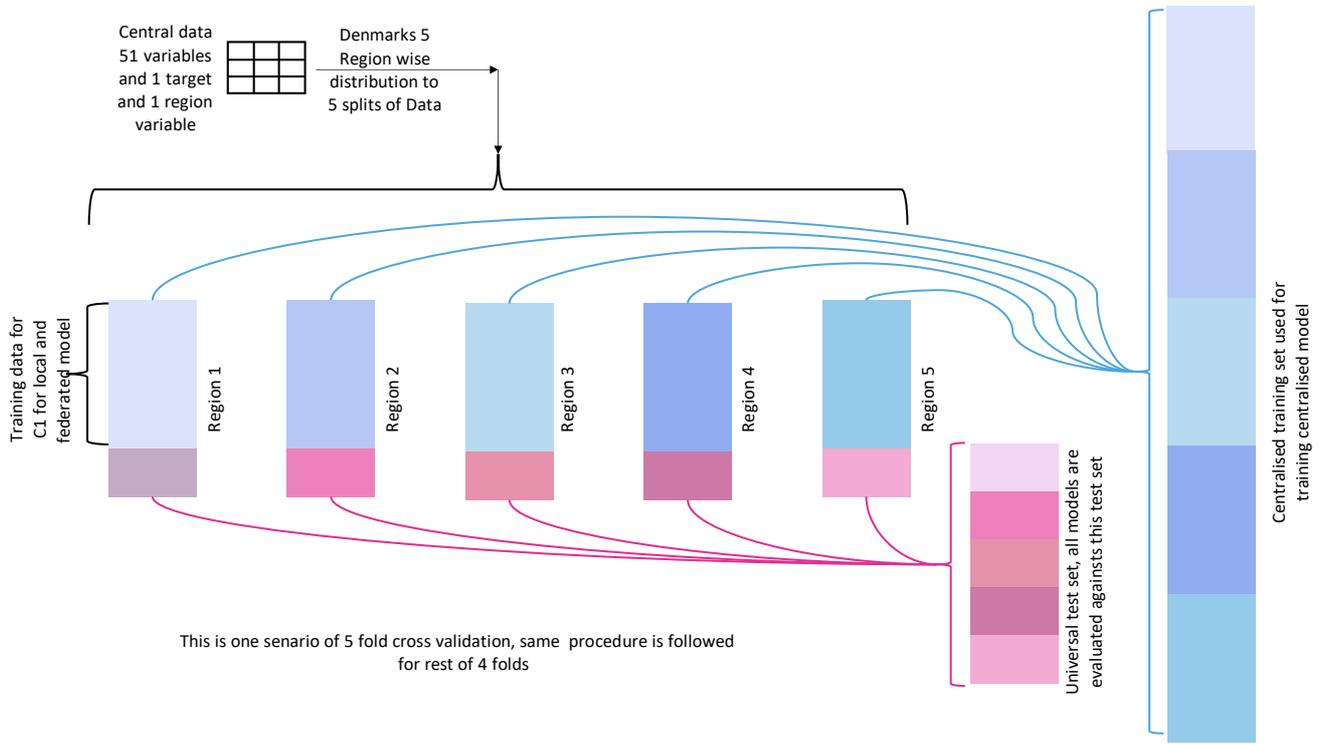

Supplementary Figure A.2: **Fair evaluation scheme**: For a fair comparison of local models with centralized and federated models and to simulate a clinically relevant scenario, we designed the same test set for all in each fold of 5-fold cross-validation. 20% of each local data was taken out (in pink) and combined to create this global identical test set. For training, each region used the rest of their data (in blue) or federated and local training, while for centralized training, a set combination of all those training data was used.

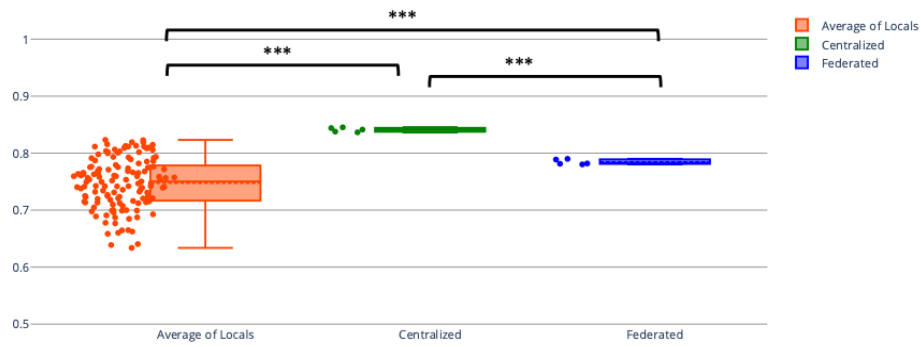

A. Random Forest Classifier – Accuracy

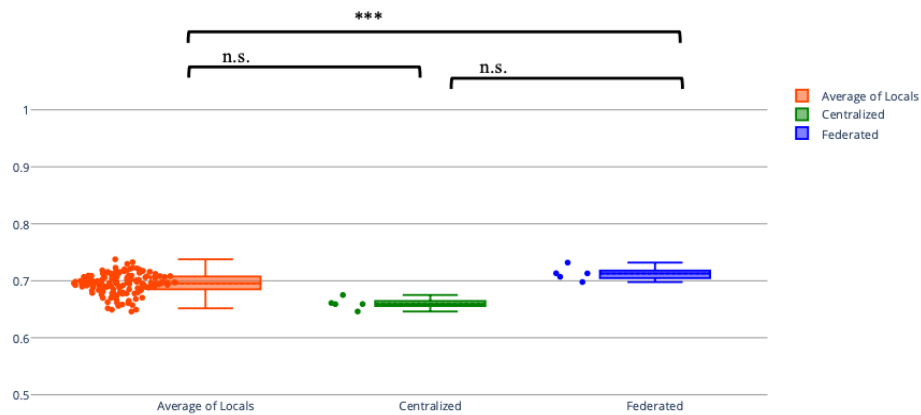

B. Random Forest Classifier – AUROC

Supplementary Figure A.3. **Comparison of federated model with centralized and local models in SHARE**: In **A.**, the random forest classifier accuracy in the centralized model is higher than that of the federated, and certain local models seem to be as good as the federated model. However, the mean (dotted line) for the average of locals is much lower because most of the local models are underperforming. The federated learning has much less variance than the local model. In real life, finding a local model that performs good in centralized data is not possible either. In **B.**, the random forest classifier AUROC in the federated model is even better than the centralized model. This suggests that federated learning is more robust. The dotted lines in the box plots represent the mean, while the solid lines represent the median. The stars represent the magnitude of the significance of the tested hypothesis in the following convention *** : p<0·001, **: p<0·01, * : p<0·05, n.s. : p>0·05. We tested the following hypotheses: 1. if the centralized model is better than the average of local ones, 2. if the federated is better than the average of local ones, and 3. if the centralized is better than the federated.

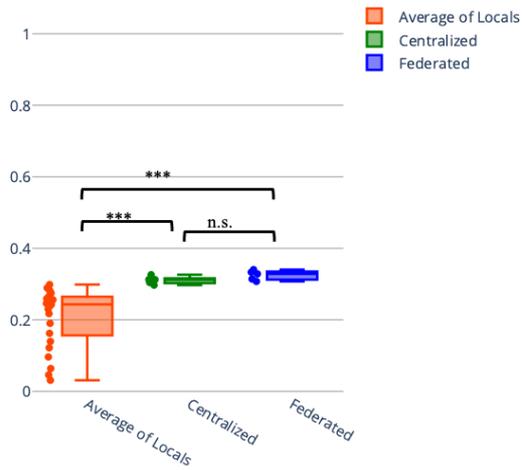

A. Linear Regression – r- square

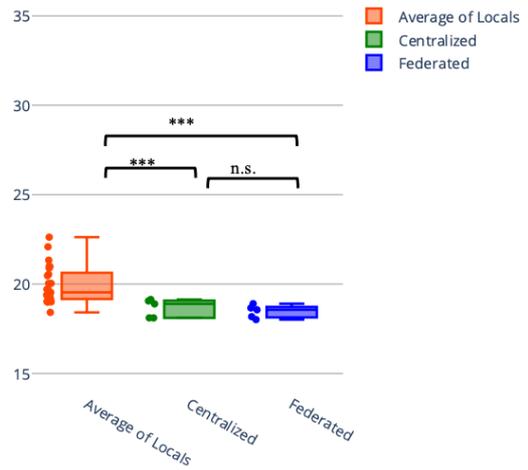

B. Linear Regression – RMSE

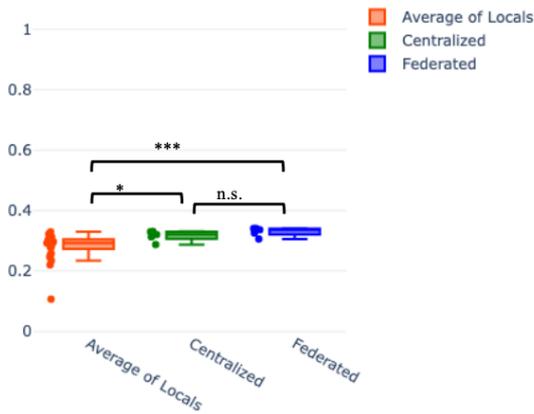

C. Random Forest Regression – r- square

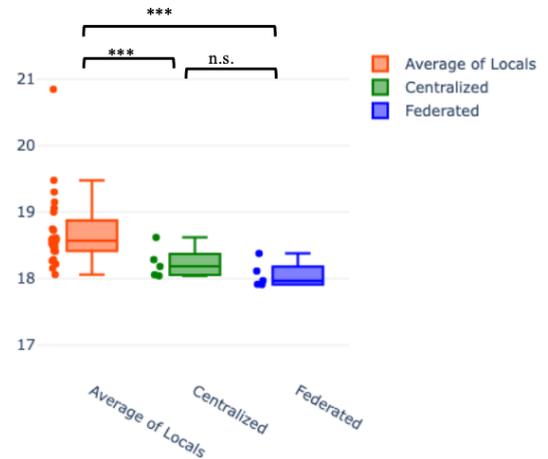

D. Random Forest Regression – RMSE

Supplementary Figure A.3. **Comparison of federated model with centralized and local models for 25% subsampled training data in *GLA:D®*:** For both **A.** R-square of linear regression (the higher, the better) and **B.** For the RMSE of linear regression (lower, better), we see that federated and centralized models perform better than the local models. Similarly, for both **C.** R-square of random forest regression (higher the better) and **D.** RMSE of random forest regression (lower the better), we see that the federated model outperforms both local and central models. The dotted lines in the box plots represent the mean, while the solid lines represent the median. The stars represent the magnitude of the significance of the tested hypothesis in the following convention *** : $p<0·001$, ** : $p<0·01$, * : $p<0·05$, n.s. : $p>0·05$. We tested the following hypotheses: 1. if the centralized model is better than the average of local ones, 2. if the federated is better than the average of local ones, and 3. if the centralized is better than the federated.

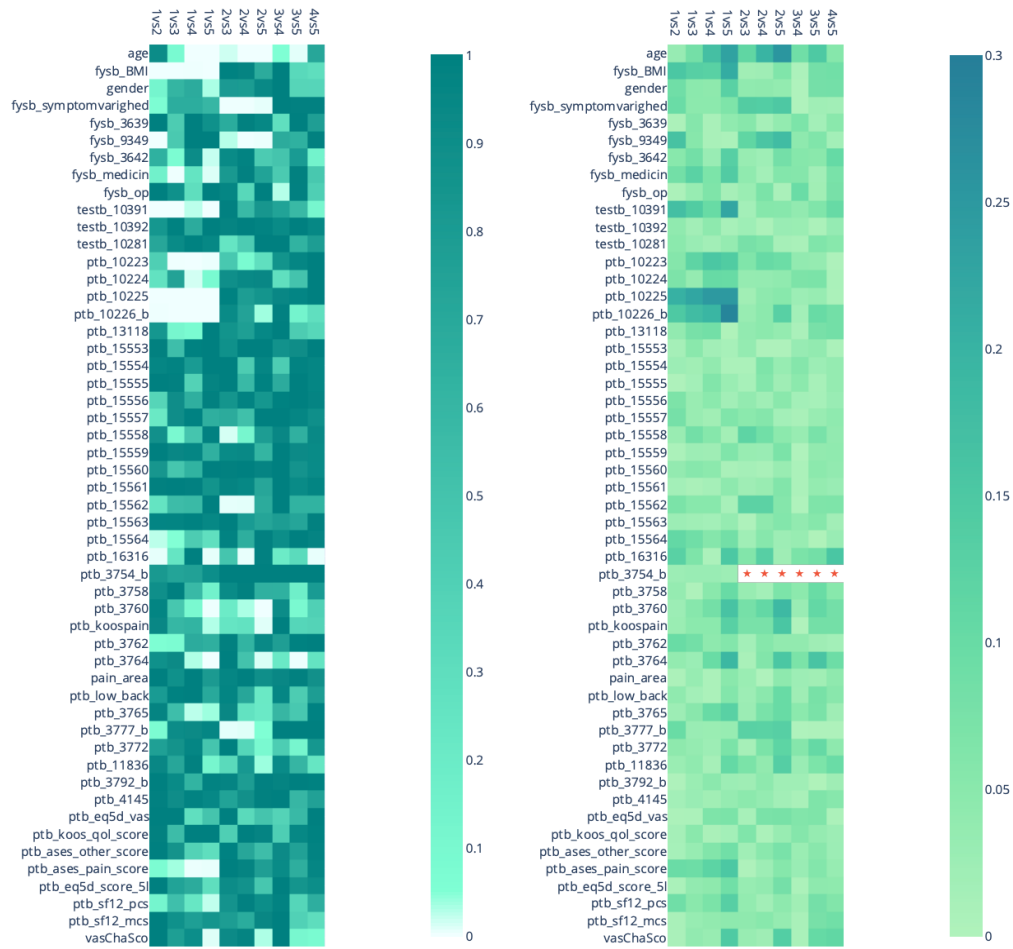

A.   Scheffe's Post-hoc                    B.   Cohen's D

Supplementary Figure A.5: **Statistical test to evaluate data heterogeneity in *GLA:D®* data**: **A. Posthoc Scheffe test:** We used post hoc Scheffe to test pairwise if two regions' data originates from different distributions. The comparison of two regions, X and Y, is denoted as "XvsY." We did this for all 51 variables and the target variable. A light color suggests statistical significance, while green suggests no significance.
**B. Cohen's-D:** This shows the Cohen's -D effect size of each statistical test; the darker the color, the higher the effect size. Red stars suggest missing values. Although the means of the distribution are close, we find statistically significant differences among regions in some variables, including the target variable.

## Average Physical Activity by Country

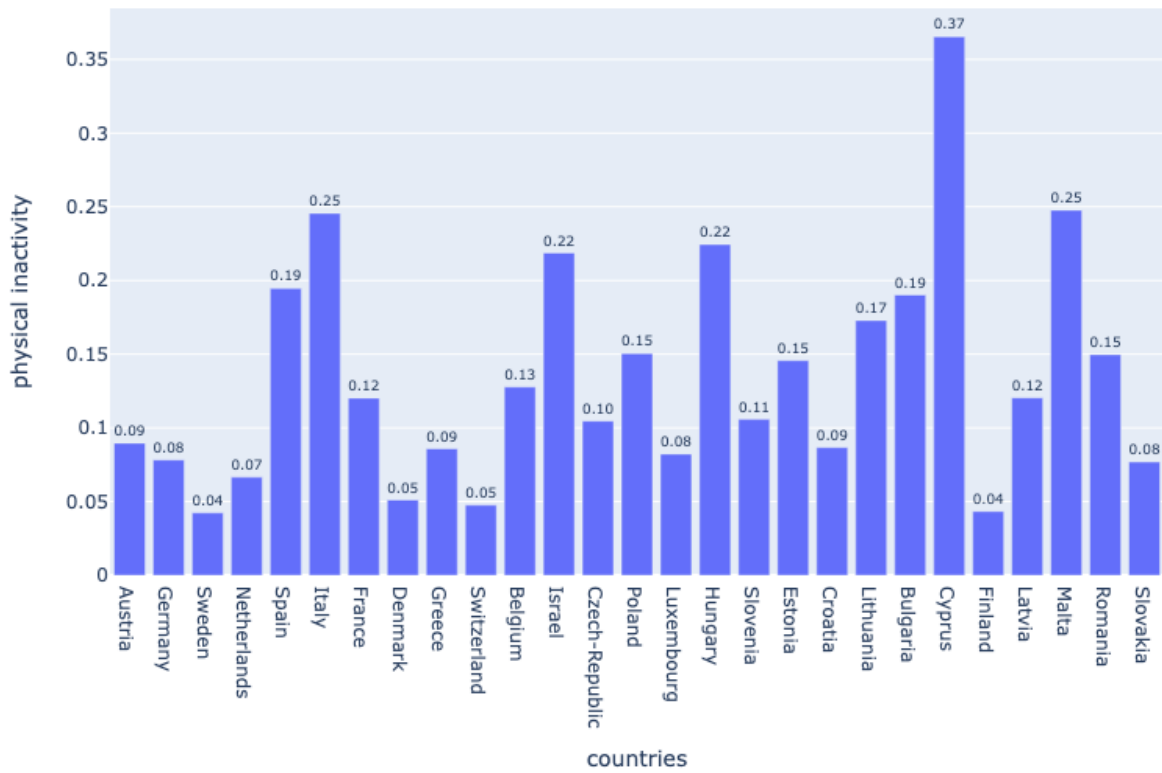

Supplementary Figure A.6: **Distribution of the target variable physical inactivity across 27 countries in SHARE data:** In the y-axis, we plot the average level of physical inactivity in each country.

**Supplementary Tables**

| Supplementary Table B.1: **GLA:D® data variable description.** | | |
|---|---|---|
| **Variable ID** | **Variable Name** | **Mean, standard deviation, (SD) or number of cases**<br>**(total n =9648)** |
| age | 1. Age | Mean:64·93<br>SD:9·40<br>Min:23<br>Max:94 |
| fysb–BMI | 2. BMI | Mean:28·60<br>SD:5·27<br>Min:15·23<br>Max:70·03 |
| gender | 3. Gender | Men:2725<br>Women:6923 |
| fysb–symptom<br>varighed | 4. Duration of symptoms | Mean:43·91<br>SD:67·76<br>Min:0·0<br>Max:756·0 |
| fysb–3639 | 5. Waitlisted for surgery | Yes:166<br>No:9482 |
| fysb–9349 | 6. Radiographic signs of knee OA | Yes:7639<br>No:373<br>Unknown:1636 |
| fysb–3642 | 7. Prior physiotherapy due to index joint | Yes:3241<br>No:6407 |
| fysb–medicin | 8. Use of painkillers (paracetamol/ NSAID/ opioids/ codeine) in the last three months | Yes:6123<br>No:3525 |
| fysb–op | 9. Earlier surgery in the index joint | Yes:2681<br>No:6967 |
| testb–10391 | 10. Time to complete 40m walking test | Mean:28·18<br>SD:7·48<br>Min:10·0<br>Max:234·91 |
| testb–10392 | 11. Use of walking aid | Yes:159<br>No:9489 |
| testb–10281 | 12. Number of chair stands during 30sec | Mean:11·95<br>SD:3·67<br>Min:0·0<br>Max:33·0 |
| ptb–10223 | 13. Born in Denmark | Yes:9287<br>No:361 |
| ptb–10224 | 14. Danish citizen | Yes:9490<br>No:158 |
| ptb–10225 | 15. Are you living alone or with others? | Living aolne:2347<br>Living with others:7301 |
| ptb–10226-b | 16. Educational level<br>(Do you have an education higher than secondary education? ) | Yes:6863<br>No:2785 |

Supplementary Table B.1: **GLA:D® data variable description.**

| Variable ID | Variable Name | Mean, standard deviation, (SD) or number of cases (total n =9648) |
|---|---|---|
| ptb-13118 | 17. Smoking | Yes:755<br>No:8893 |
| ptb-15553 | 18. Hypertension | Yes:3652<br>No:5996 |
| ptb-15554 | 19. Heart disease | Yes:699<br>No:8949 |
| ptb-15555 | 20. Lung disease e.g. COpD | Yes:552<br>No:9096 |
| ptb-15556 | 21. Diabetes | Yes:675<br>No:8973 |
| ptb-15557 | 22. Digestion disease | Yes:445<br>No:9203 |
| ptb-15558 | 23. Liver or kidney disease | Yes:113<br>No:9535 |
| ptb-15559 | 24. Blood disease e.g. Anemia | Yes:107<br>No:9541 |
| ptb-15560 | 25. Cancer | Yes:254<br>No:9394 |
| ptb-15561 | 26. Depression | Yes:365<br>No:9283 |
| ptb-15562 | 27. Rheumatoid arthritis | Yes:468<br>No:9180 |
| ptb-15563 | 28. Neurological disease e.g. Parkinson | Yes:457<br>No:9191 |
| ptb-15564 | 29. Other medical disease | Yes:1743<br>No:7905 |
| ptb-16316 | 30. Previous injury in the index joint serious enough to consult a medical doctor | Yes:4951<br>No:4697 |
| ptb-3754-b | 31. Pain in the hip or knee other than the index joint? | Yes:1<br>No:9647 |
| ptb-3758 | 32. Walking difficulty | Yes:7397<br>No:2251 |
| ptb-3760 | 33. Pain and problems with hands and fingers | Yes:3229<br>No:6419 |
| ptb-koospain | 34. Knee pain at least every day? | Yes:7782<br>No:1866 |
| ptb-3762 | 35. Anxious about physical activity | Yes:1429<br>No:8219 |
| ptb-3764 | 36. Baseline pain intensity during the last month (VAS scale 0-100, no pain to worst pain | Mean:46·76<br>SD:21·68<br>Min:0·0<br>Max:100·0 |
| pain-area | 37. Painful body areas (collected via pain drawing) | Mean:3·45<br>SD:3·26<br>Min:0·0<br>Max:40·0 |
| ptb-low-back | 38. Pain in the lower back (collected via pain drawing) | Yes:2334<br>No:7314 |
| ptb-3765 | 39. Wants to undergo surgery | Yes:1037<br>No:8611 |

Supplementary Table B.1: **GLA:D® data variable description.**                    OA:

| Variable ID | Variable Name | Mean, standard deviation, (SD) or number of cases (total n =9648) |
|---|---|---|
| ptb-3777-b | 40. Working status<br>Are you working/ studying? | Working/Studing:2875<br>No:6773 |
| ptb-3772 | 41. Sick leave during the last year because of knee/hip during last year? | Yes:1085<br>No:8563 |
| ptb-11836 | 42. Replaced knee or hip joints | Yes:877<br>No:8771 |
| ptb-3792-b | 43. Frequency of training until exhaustion at least 2-3 times a week? | Yes:4913<br>No:4735 |
| ptb-4145 | 44. UCLA - physical activity score<br>(from 0 : 10 worst to best) | Mean:5·76<br>SD:1·79<br>Min:1·0<br>Max:10·0 |
| ptb-eq5d-vas | 45. Health situation evaluated via VAS<br>(from 0 : 100, worst to best) | Mean:70·43<br>SD:18·41<br>Min:0·0<br>Max:100·0 |
| ptb-koos-qol-score | 46. KOOS quality of life score<br>(from 0 : 100, worst to best) | Mean:45·79<br>SD:15·06<br>Min:0·0<br>Max:100·0 |
| ptb-ases-pain-score | 48. ASES pain score<br>(from 10 : 100, worst to best) | Mean:67·63<br>SD:19·26<br>Min:10·0<br>Max:100·0 |
| ptb-ases-other-score | 47. ASES other score (from 10 : 100, worst to best) | Mean:71·54<br>SD:17·20<br>Min:10·0<br>Max:100·0 |
| ptb-eq5d-score-5l | 49. EQ-5D score<br>From -0.624 : 1, worst to best) | Mean:0·78<br>SD:0·18<br>Min:-0·37<br>Max:1·0 |
| ptb-sf12-pcs | 50. SF12 score physical component<br>(from 0 : 100, worst to best) | Mean:38·27<br>SD:8·85<br>Min:11·03<br>Max:65·95 |
| ptb-sf12-mcs | 51. SF12 score mental component<br>(from 0 : 100, worst to best) | Mean:53·03<br>SD:9·43<br>Min:11·62<br>Max:70·99 |
| vasChaSco | 52. Improvement in VAS pain score from baseline to 3 months follow-up score<br>-100 : 100, from pain got worse to pain got better | Mean:14·23<br>SD:22·48<br>Min:-87·0<br>Max:99·0 |

Osteoarthritis, NSAID: non-steroidal anti-inflammatory drugs, VAS: visual analog scale, KOOS: Knee Injury and Osteoarthritis Outcome Score, ASES:  Arthritis Self Efficacy Scale, EQ-5D: EuroQol- 5 dimensions.SF-12" 12-Item Short Form Survey.

Supplementary Table B.2: **Sample size distribution across 27 countries in SHARE data.**

| Country | Patients | Country | Patients | Country | Patients |
|---|---|---|---|---|---|
| Austria | 1483 | Switzerland | 1869 | Croatia | 1153 |
| Germany | 2797 | Belgium | 1941 | Lithuania | 1411 |
| Sweden | 2286 | Israel | 809 | Bulgaria | 899 |
| Netherlands | 1879 | Czech-Republic | 2599 | Cyprus | 509 |
| Spain | 1952 | Poland | 1986 | Finland | 1108 |
| Italy | 2044 | Luxembourg | 863 | Latvia | 772 |
| France | 2403 | Hungary | 753 | Malta | 763 |
| Denmark | 2130 | Slovenia | 2353 | Romania | 1269 |
| Greece | 2933 | Estonia | 2861 | Slovakia | 988 |

| Supplementary Table B.3: **SHARE data variable description.** | | |
|---|---|---|
| **Variable id** | **Variable name** | **Mean, standard deviation, (SD) or number of cases (total n =44813)** |
| bmi | 1. BMI | Mean: 27·17<br>SD: 4·73<br>Min: 12·46<br>Max: 97·78<br>Don't know/ Refusal to answer/suspected wrong: 1205 |
| sphus | 2. Self-perceived health (US version)<br>(Based on SF-36 questionnaire )<br>(from 1 : 5, excellent to poor) | Mean: 3·19<br>SD: 1·0<br>Min: 1<br>Max: 5<br>Don't know/ Refusal to answer: 12 |
| sphus2 | 3. Do you have less than very good health in self-perceived health (sphus)? | Yes: 34689<br>No: 10112<br>Don't know/ Refusal to answer: 12 |
| chronic2w8 | 4. Do you have 2+ chronic diseases? | Yes: 23680<br>No: 21102<br>Don't know/ Refusal to answer: 31 |

| Supplementary Table B.3: **SHARE data variable description.** | | |
|---|---|---|
| **Variable id** | **Variable name** | **Mean, standard deviation, (SD) or number of cases** <br> **(total n =44813)** |
| chronicw8c | 5. Number of chronic diseases | Mean: 1·90 <br> SD: 1·60 <br> Min: 0 <br> Max: 14 <br> Don't know/ Refusal to answer: 31 |
| numeracy2 | 6. Score of second numeracy test <br> (from 0 : 5, bad to good) | Mean: 4·11 <br> SD: 1·41 <br> Min: 0 <br> Max: 5 <br> Don't know/ Refusal to answer: 0 |
| orienti | 7. Score of orientation in time test <br> (from 0 : 4, bad to good) | Mean: 3·83 <br> SD: 0·53 <br> Min: 0 <br> Max: 4 <br> Don't know/ Refusal to answer: 0 |
| cf008tot | 8. Ten words list learning the first trial | Mean: 5.27 <br> SD: 1.75 <br> Min: 0 <br> Max:10 |
| cf016tot | 9. Ten words list learning delayed recall | Mean: 3·87 <br> SD: 2·14 <br> Min: 0 <br> Max: 10 |
| euro1 | 10. Depression <br> In the last month, have you been sad or depressed? <br> (part of EURO-D measure of depressive Symptoms) | Yes: 17775 <br> No: 26985 <br> Don't know/ Refusal to answer: 53 |
| euro2 | 11. Pessimism <br> What are your hopes for the future? <br> (part of EURO-D measure of depressive Symptoms) | Any hopes mentioned: 7895 <br> No such mentioned: 36815 <br> Don't know/ Refusal to answer: 103 |
| euro3 | 12. Suicidality <br> In the last month, have you felt that you would rather be dead? <br> (part of EURO-D measure of depressive Symptoms) | Any such mention: 2634 <br> No such mention: 42093 <br> Don't know/ Refusal to answer: 86 |
| euro4 | 13. Guilt <br> Do you tend to blame yourself or feel guilty about anything? <br> (part of EURO-D measure of depressive Symptoms) | Obvious mention of guilt / self-blame: 3370 <br> No: 41366 <br> Don't know/ Refusal to answer: 77 |
| euro5 | 14. Sleep <br> Have you had trouble sleeping recently? <br> (part of EURO-D measure of depressive Symptoms) | Yes: 16362 <br> No: 28428 <br> Don't know/ Refusal to answer: 23 |
| euro6 | 15. Interest <br> In the last month, what is your interest in things? <br> (part of EURO-D measure of depressive Symptoms) | Less interest than normal: 4771 <br> No mention of the loss of interest: 40007 <br> Don't know/ Refusal to answer: 35 |

| Supplementary Table B.3: **SHARE data variable description.** | | |
|---|---|---|
| **Variable id** | **Variable name** | **Mean, standard deviation, (SD) or number of cases (total n =44813)** |
| euro7 | 16. Irritability<br>Have you been irritable recently?<br>(part of EURO-D measure of depressive Symptoms) | Yes: 33211<br>No: 11558<br>Don't know/ Refusal to answer: 44 |
| euro8 | 17. Appetite<br>What has your appetite been like?<br>(part of EURO-D measure of depressive Symptoms) | Diminution in desire for food: 4196<br>No diminution: 40606<br>Don't know/ Refusal to answer: 11 |
| euro9 | 18. Fatigue<br>In the last month, have you had too little energy to do things you wanted to do?<br>(part of EURO-D measure of depressive Symptoms) | Yes: 15748<br>No: 29020<br>Don't know/ Refusal to answer: 45 |
| euro10 | 19. Concentration<br>How is your concentration?<br>(part of EURO-D measure of depressive Symptoms) | Difficulty in concentration: 7574<br>No difficulty: 37215<br>Don't know/ Refusal to answer: 24 |
| euro11 | 20. Enjoyment<br>What have you enjoyed doing recently?<br>(part of EURO-D measure of depressive Symptoms) | Fails to mention any: 6027<br>Mentioned any activity: 38722<br>Don't know/ Refusal to answer: 64 |
| euro12 | 21. Tearfulness<br>In the last month, have you cried at all?<br>(part of EURO-D measure of depressive Symptoms) | Yes: 10617<br>No: 34155<br>Don't know/ Refusal to answer: 41 |
| loneliness | 22. Loneliness (short version of R-UCLA Loneliness Scale)<br>(from 3: 9, not lonely to very lonely) | Mean: 3·96<br>SD: 1·41<br>Min: 3<br>Max: 9<br>Don't know/ Refusal to answer: 0 |
| gali | 23. Limitations with activities | Yes: 21660<br>No: 23137<br>Don't know/ Refusal to answer: 16 |
| mobility | 24. Number of limitations in mobility, arm function, and fine motor limitations | Mean: 1·71<br>SD: 2·37<br>Min: 0<br>Max: 10<br>Don't know/ Refusal to answer: 14 |
| mobilit2 | 25. Do you have 1+ mobility, arm function, and fine motor limitations? | Yes: 23069<br>No: 21730<br>Don't know/ Refusal to answer: 14 |
| mobilit3 | 26. Do you have 3+ mobility, arm function, and fine motor limitations? | Yes: 11983<br>No: 32816<br>Don't know/ Refusal to answer: 14 |
| adl | 27. Number of limitations with activities of daily living (ADL) | Mean: 0·22<br>SD: 0·81<br>Min: 0<br>Max: 6<br>Don't know/ Refusal to answer: 20 |
| adl2 | 28. Do you have 1+ ADL limitations? | Yes: 4910<br>No: 39883<br>Don't know/ Refusal to answer: 20 |

| Supplementary Table B.3: **SHARE data variable description.** | | |
|---|---|---|
| **Variable id** | **Variable name** | **Mean, standard deviation, (SD) or number of cases (total n =44813)** |
| iadl | 29. Number of limitations with instrumental activities of daily living | Mean: 0·50 SD: 1·42 Min: 0 Max: 9 Don't know/ Refusal to answer: 20 |
| iadl2 | 30. Do you have 1+ IADL limitations? | Yes: 8238 No: 36555 Don't know/ Refusal to answer: 20 |
| inact | 31. Calculated physical inactivity | Yes: 8238 No: 36575 |

sphus = Self-Perceived Health US version, Abbreviations used: ADL = Activities of daily living, IADL = Limitations with instrumental activities of daily living